\documentclass{article}
\usepackage{spconf,amsmath,graphicx}
\usepackage{booktabs}
\usepackage[colorlinks,
linkcolor=black,
citecolor=black,
anchorcolor=black,
urlcolor=black]{hyperref}
\title{Abnormal Respiratory Patterns Classifier may contribute to large-scale screening of people infected with COVID-19 in an Accurate and Unobtrusive manner}
\name{Yunlu Wang$^{1}$, Menghan Hu$^{1}$, Qingli Li$^{1}$, Xiao-Ping Zhang$^{3}$, Guangtao Zhai$^{2}$,Nan Yao$^{4}$}
\address{$^1$Shanghai Key Labora. of Multidim. Infor. Proce., East China Normal University, China\\
$^2$Key Laboratory of Artiﬁcial Intelligence, Ministry of Education, China\\
$^3$Department of Electrical, Computer and Biomedical Engineering, Ryerson University, Canada\\
$^4$ Shanghai Jianglai Data Technology Co., Ltd, China
\thanks{This work is sponsored by the Shanghai Sailing Program (No.19YF1414100), the National Natural Science Foundation of China (No.61901172), the STCSM (No.18DZ2270700), and the Science and Technology Commission of Shanghai Municipality (No.19511120100).}\\
\thanks{Corresponding author: Menghan Hu (mhhu@ce.ecnu.edu.cn)}}
\begin{document}
%
\maketitle

\begin{abstract}
\textbf{Research significance}: The extended version of this paper has been accepted by IEEE Internet of Things journal (DOI: 10.1109/JIOT.2020.2991456), please cite the journal version. During the epidemic prevention and control period, our study can be helpful in prognosis, diagnosis and screening for the patients infected with COVID-19 (the novel coronavirus) based on breathing characteristics. According to the latest clinical research, the respiratory pattern of COVID-19 is different from the respiratory patterns of flu and the common cold. One significant symptom that occurs in the COVID-19 is Tachypnea. People infected with COVID-19 have more rapid respiration. Our study can be utilized to distinguish various respiratory patterns and our device can be preliminarily put to practical use. Demo videos of this method working in situations of one subject and two subjects can be downloaded online (\url{https://doi.org/10.6084/m9.figshare.11493666.v1}).\\
\textbf{Research details}: Accurate detection of the unexpected abnormal respiratory pattern of people in a remote and unobtrusive manner has great significance. In this work, we innovatively capitalize on depth camera and deep learning to achieve this goal. The challenges in this task are twofold: the amount of real-world data is not enough for training to get the deep model; and the intra-class variation of different types of respiratory patterns is large and the outer-class variation is small. In this paper, considering the characteristics of actual respiratory signals, a novel and efficient Respiratory Simulation Model (RSM) is first proposed to fill the gap between the large amount of training data and scarce real-world data. Subsequently, we first apply a GRU neural network with bidirectional and attentional mechanisms (BI-AT-GRU) to classify 6 clinically significant respiratory patterns (Eupnea, Tachypnea, Bradypnea, Biots, Cheyne-Stokes and Central-Apnea). The performance of the obtained BI-AT-GRU is tested by real-world data measured by depth camera, and the results show that the proposed model can classify 6 different respiratory patterns with the accuracy, precision, recall and F1 of 94.5\%, 94.4\%, 95.1\% and 94.8\% respectively. In comparative experiments, the obtained BI-AT-GRU specific to respiratory pattern classification outperforms the existing state-of-the-art models. The proposed deep model and the modeling ideas have the great potential to be extended to large scale applications such as public places, sleep scenario, and office environment.
\end{abstract}
\section{Introduction}
\label{sec:intro}
Respiration is a core physiological process for all living creatures on earth. A person’s physiological state \cite{griffiths2019guidelines} as well as emotion \cite{hameed2019human} and stress \cite{perciavalle2017role} may be reflected by representation of some respiratory parameters. Therefore, we should pay attention to respiration. Integrated various respiratory signs, respiratory patterns are able to more comprehensively reflect the conditions of respiratory activity. Many clinical literatures suggested that abnormal respiratory patterns are able to predict a few specific diseases \cite{holfinger20190459}, providing relatively detailed clues for clinical treatments \cite{boulding2016dysfunctional,koyama2019application}. Unfortunately, these abnormal respiratory patterns occur in a way difficult for people to notice themselves. If we could develop a system capable of remotely and unobtrusively detecting these unnoticeable abnormal breathing patterns under various scenarios, people who have the diseases may be diagnosed at earliest possible time.

Contact measurement devices are heavy, expensive, and inconvenient for patients to use \cite{al2011respiration}. Therefore, non-contact measurement methods are more suitable for detecting abnormal respiratory patterns. Doppler radar, thermal imaging technology and camera based on motion detection are usually used for recording respiratory signals. The continual use of radar has the potential risk of the explosion to released radiation \cite{kranjec2013novel}. Thermal imaging cameras are susceptible to ambient heat \cite{al2017monitoring,hu2018combination}. The method based on motion detection is able to solve the above problems, because its measurement is based on the variation in displacement instead of the changes in pixel intensity. Depth camera is a good method based on motion detection which has been widely utilized in the acquisition and analysis of respiratory signals \cite{soleimani2016remote,bae2018development}. Nonetheless, most relevant work focused on improving the accuracy of extracting respiratory signals and restraining the noise in movement \cite{ernst2015respiratory}. Some paid attention to the detection of abnormal respiratory patterns, but did not classify them specifically \cite{al2017real}. Thus, the available clinical clues were limited.

Deep learning has been extensively used in many fields. In terms of respiratory pattern detection, Cho et al. capitalized on the convolutional neural networks (CNN) to achieve the identification of deep breathing \cite{cho2017deepbreath}. Kim et al. used 1D CNN to classify the respiratory signals estimated by radars into four categories \cite{kim20191d}. Therefore, the classification of respiratory signals extracted by the non-contact measurement system with the aid of deep learning is a study worth trying and of much significance. In the above study, all the data sets required for model construction were obtained, though, by assessing the respiratory activities of the test subjects. This approach for capturing different types of respiratory patterns yields a limited set of data. Were there no large amount of training data to support, the neural network might not be able to give full play to its advantages. Furthermore, researchers often present studies that the classification models generally adopt the general network architecture in the field of deep learning without specific designs for respiratory pattern classification. At the same time, the network is not optimized according to the characteristics of the collected data.

In this paper, we pioneer an AT-BI-GRU deep neural network for classifying abnormal respiratory patterns. To fill the gap between the large amount of training data and scarce real-world data, we first propose Respiratory Simulation Model (RSM) to generate abundant reliable data. The obtained classifier validated by 605 real-world data measured by depth camera can be used to achieve the remote and unobtrusive measurement of respiratory patterns.

\section{MODELLING PROCEDURE}
\label{sec:method}

The	proposed respiratory pattern classification model consists of four parts: 1) develop respiratory simulation model for generating simulated data; 2) acquire real-world data using depth camera; 3) establish and validate BI-AT-GRU model; and 4) conduct the comparative experiments.

\subsection{Respiratory patterns}
\label{ssec:subhead2.1}
Respiratory patterns can be divided into normal breathing (Eupnea) and abnormal breathing such as Bradypnea, Tachypnea, Biots, Cheyne-Stokes and Central-Apnea. Standard waveforms are available at the website (\url{https://doi.org/10.6084/m9.figshare.9981236.v2}).

\subsection{Respiratory Simulation Model (RSM)}
\label{ssec:subhead2.2}

Respiration is a cyclic process of inhalation and exhalation, which are reflected in the rise and fall of the waveform. Thus, respiratory signals measured by non-contact method can be approximated by sine wave. Actual measured respiratory signals, especially those measured by non-contact method, are prone to deviation due to environmental changes, leading to fluctuations in respiratory depth and frequency within a certain range. Affected by body movements in measurements, signals also tend to have longitudinal and oblique deviation. Considering the possible deviation mentioned above, the actual measured respiratory signals can be defined by the following equation:
$$y_{i\in\omega}=a_{i}\sin(b_{i}x)+c_i+d_{i}x\eqno{(1)}$$
where $\omega$ is a constant time window, indicating the cycle of examining a respiratory pattern; $i$ is a variable time, which is used to represent changes of signals belonging to the same respiratory pattern; the point dividing each period of $i$ can be called a ‘breakpoint’; $a_{i}$ is respiratory depth; ${b}_i$ is respiratory rate; $c_{i}$ and $d_{i}$ are longitudinal deviation degree and oblique deviation degree of respiratory signal, respectively. Gaussian white noise is added to the simulated respiratory signal to make it closer to the real-world measured respiratory signal.

By changing values of $a_i$, $b_i$, $c_i$ and $d_i$ in equation (1), different features in the process of respiration can be simulated, such as speed, depth and time of apnea. A specific partial waveform can be generated by randomizing the parameters of each waveform in a pre-set range. We combine every partial waveform through breakpoints then specific respiratory patterns can be obtained. The random range of parameters is set for the 6 respiratory patterns by reference to standard waveforms and real-world waveforms, and it can be acquired in our released RSM \href{https://doi.org/10.6084/m9.figshare.9978833.v1}{code} (\url{https://doi.org/10.6084/m9.figshare.9978833.v1}). To illustrate the process of generating data, we select an abnormal respiratory pattern to demonstrate (\url{https://doi.org/10.6084/m9.figshare.9981311.v1}).
\subsection{Measuring respiratory signal by depth camera}
\label{ssec:subhead2.3}

The depth camera is used to conduct non-contact respiratory signal measurement to obtain real-world data. Our measurement process includes obtaining depth images via the depth camera, selecting region of interest (ROI) and processing depth data.

A Kinect v2 depth camera was used to record depth images of subjects when they breathed. A total of 20 participants were asked to sit in a chair and learn to imitate 6 respiratory patterns. Spirometer of a sleep monitor (GY-6620, HeNan HuaNan Medical Science and Technology Co., LTD.) was used to check their respiratory situations to make sure they had learned the pattern. If real-world respiratory data is inconsistent with these obtained by the gold standard method, we excluded them for subsequently modeling. Subjects were at 1-4m from the depth camera, and were asked to breathe in a specific respiratory pattern for one minute at a time. The recording frame number of Kinect v2 was set as 10 fps.

\begin{figure}[ht]
\centering
\includegraphics[height=5cm,width=6.5cm]{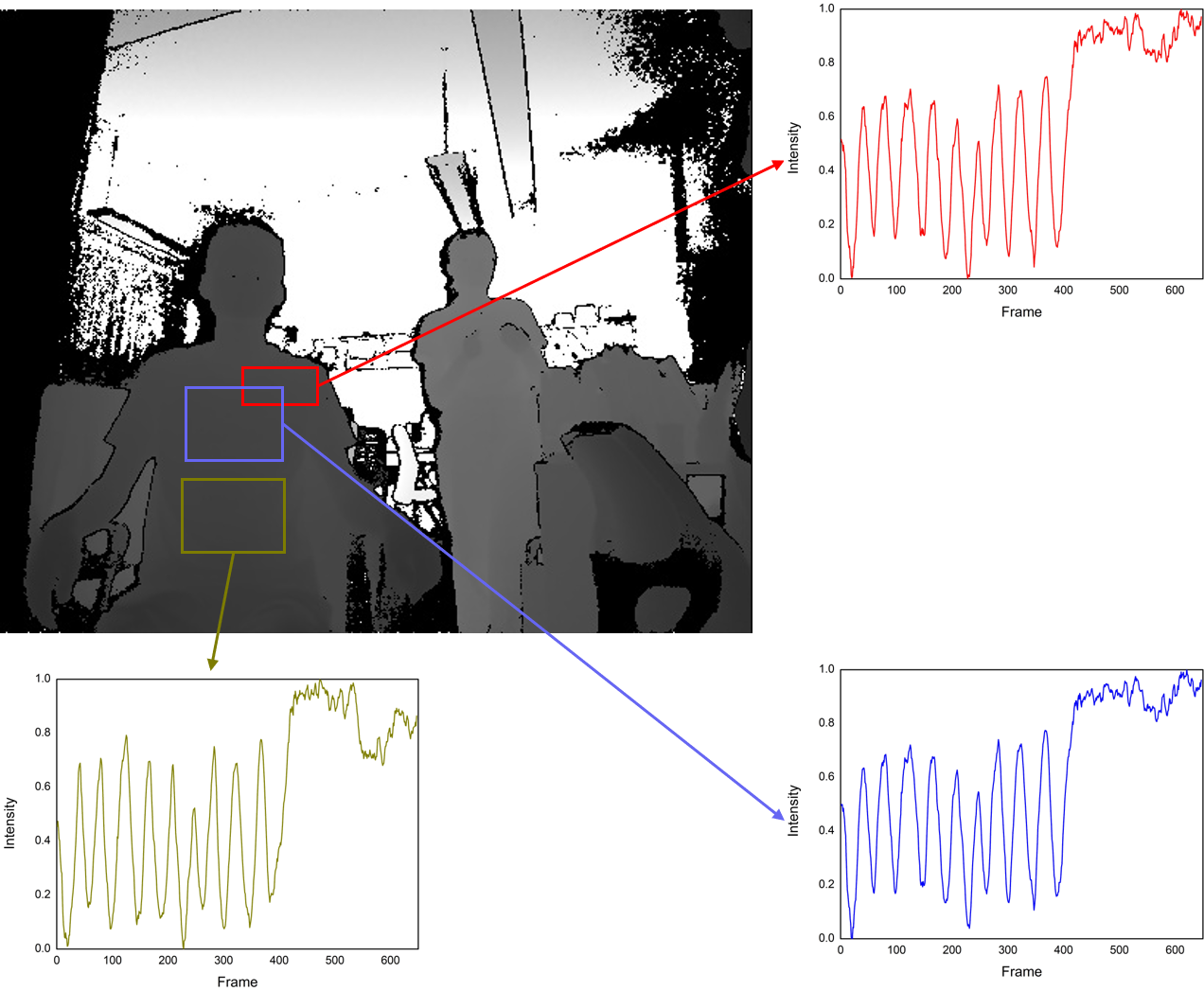}
\caption{Actual measured Central-Apnea waveforms of one subject specific to three ROIs.}
\label{Fig3}
\end{figure}

In depth images, we selected three ROIs, namely chest, abdomen and shoulder. These ROIs not only solve the problem that a single ROI may not work for some people, but also increase the amount of data obtained by one measurement. We calculated the average value of depth data in certain ROI of each frame to extract respiratory signals. Subsequently, all frames of raw data were smoothed by a moving average filter with a data span of 5 to eliminate sudden changes of waveform. Finally, min-max normalization was carried out on respiratory signals. \textbf{Fig. \ref{Fig3}} shows actual measured Central-Apnea waveforms of one subject specific to three ROIs.

\subsection{BI-AT-GRU for respiratory patterns classification}
\label{subsection2.4}

According to characteristics of our task, we first apply BI-AT-GRU to classify respiratory patterns. BI-AT-GRU was trained by simulation data generated by RSM and was tested by real-world data measured by depth camera.

Both simulated respiratory data and real-world respiratory data can be regarded as time series data, a kind of sequential data. Recurrent Neural Network (RNN) is a neural network model, and is very suitable for sequential data modeling \cite{elman1990finding}. Long-Short Term Memory (LSTM) is an important variant network of RNN, and can solve the problem of long training time of RNN and long-term memory loss in long sequences \cite{hochreiter1997long}. Gated Recurrent Unit (GRU) \cite{cho2014learning} is a simplified variant of LSTM.

\begin{figure*}[ht]
\centering
\includegraphics[height=6.33cm,width=13.65cm]{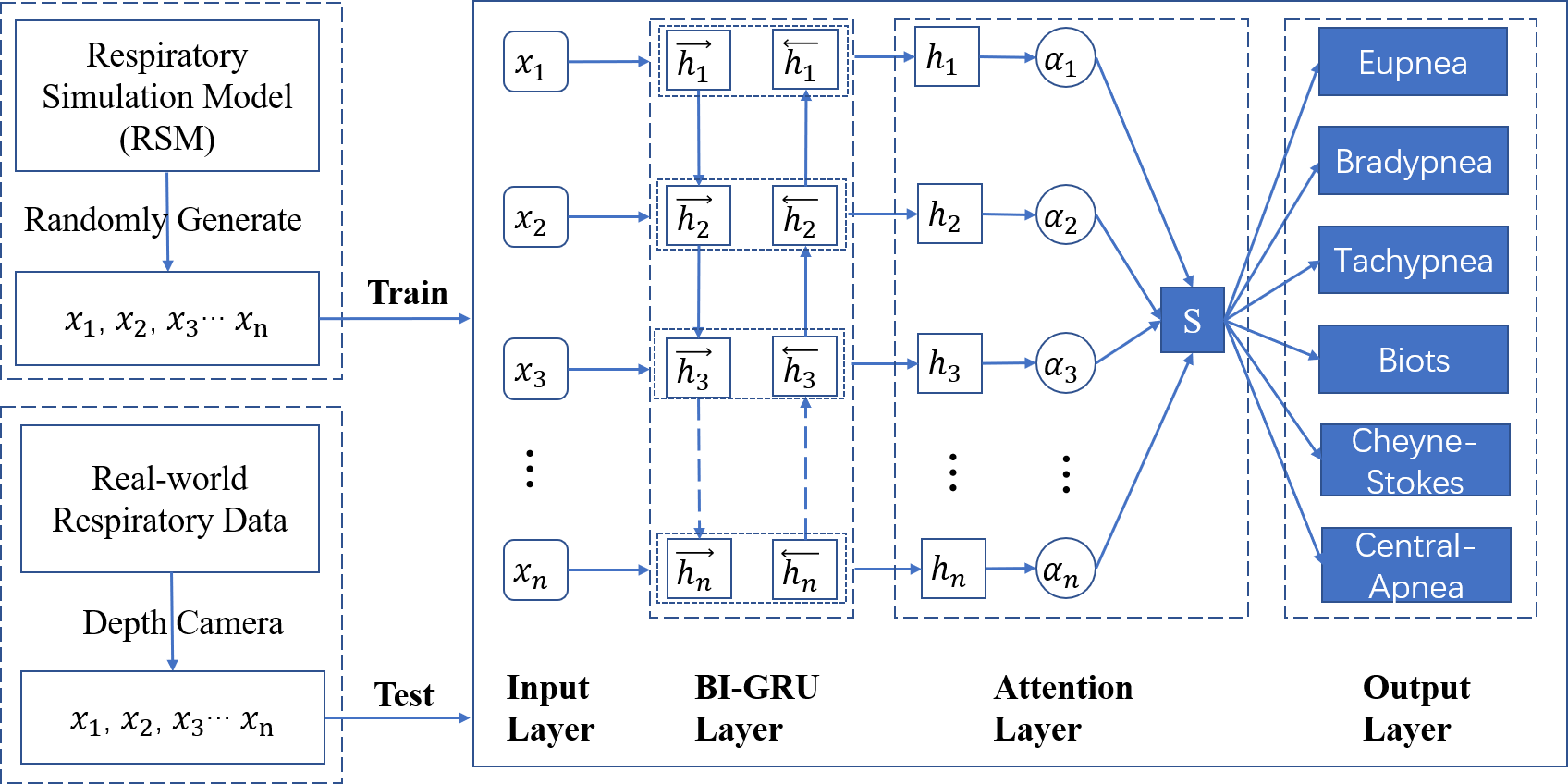}
\caption{BI-AT-GRU for respiratory pattern classification.}
\label{Fig4}
\end{figure*}

We utilized an improved GRU viz. BI-AT-GRU for respiratory pattern classification by adding bidirectional and attentional mechanisms to GRU network. The network architecture is shown in \textbf{Fig. \ref{Fig4}}. The whole network is divided into input layer, BI-GRU layer, attention layer and output layer. The input layer is used to input simulated data (training stage) or depth data (testing stage) at each point of respiratory waveform.

By observing and studying different respiratory pattern waveforms measured by depth camera, we found two particular characteristics: 1) when judging respiratory patterns, compared with left to right observation of respiratory waveform, observation in reverse chronological order (right to left observation) can acquire more information; and 2) there are key “turning points” in some combined respiratory patterns such as Central-Apnea which consists of normal breathing and apnea. Those turning points can provide reliable evidence for judging respiratory patterns. Based on these two findings, we added bidirection and attention into GRU, respectively corresponding to BI-GRU layer and attention layer in BI-AT-GRU.

\textbf{Bidirection:} Bidirectional RNN (BI-RNN) was proposed by Bahdanau et al. \cite{bahdanau2014neural} and used in text translation. Inspired by this work, we added a bidirectional mechanism to GRU to obtain forward sequence information $\overrightarrow{h_t} $ from beginning to end of breathing process and backward sequence information $\overleftarrow{h_t}$ from end to beginning. This can be expressed by following equations:
$$\overrightarrow{h_t}=\overrightarrow{GRU}(x_t),\qquad t\in{}[1,T]\eqno(2)$$
$$\overleftarrow{h_t}=\overleftarrow{GRU}(x_t), \qquad t\in{}[T,1]\eqno(3)$$
$${h_t}=[\overrightarrow{h_t},\overleftarrow{h_t}]\eqno(4)$$
where $x_{t}$ is input simulated data generated by RSM (in training stage) or input depth data (in testing stage) at time $t$; $T$ represents the period of a respiratory pattern.

\textbf{Attention:} Attention RNN has been widely applied in document classification \cite{yang2016hierarchical}, speech recognition \cite{chorowski2015attention} and relation classification \cite{zhou2016attention}. In above studies, the attention mechanism is used to express the importance of words in sentences or sentences in documents. In our task—classification of respiratory patterns, each point in the respiratory waveform is of different importance when determining certain respiratory pattern. Therefore, we add attentional mechanism into GRU, which can be expressed by following equations:
$$u_{t}=\tanh{(W_{a}h_{t}+b_{a})}\eqno{(5)}$$
$$\alpha_{t}=softmax(V_{a}u_{t})\eqno{(6)}$$
$$S=\sum_{t}\alpha_{t}h_{t}\eqno{(7)}$$
where $h_t$ is the state at time $t$, which contains bidirectional information extracted from the waveform; $W_a$ and $b_a$ are the parameters obtained by training phase. First, a $tanh$ function is used to obtain the hidden representation $u_t$. Then normalized importance weight $\alpha_t$ is obtained through a $softmax$ function. ${V_a}$ is also a parameter obtained by training processing and it can be understood as the importance of one specific point to the certain respiration pattern. Finally, the sum of product of $\alpha_t$ and $h_t$ at each point is calculated to obtain the output of the attention layer, which is denoted as $S$. $S$ here is high-level representation of certain waveform for classification. Output layer uses the output $S$ of the attention layer to classify respiratory patterns. In training stage, we encoded the label in one-hot format and cross-entropy was adopted as loss function.

\section{Experiment and results}
\label{sec3}

\subsection{Experimental settings}
\label{sub3.1}
The size of training set was 120,000 including 6 respiratory patterns. Each pattern had 20,000 samples randomly generated by the proposed RSM. Hidden layers, attention hidden layers and batch size were 128, 16, 128, respectively. To get real-world data of 6 respiratory patterns, we measured the respiratory signal of 20 subjects (12 female and 8 male). Subjects were instructed to imitate every respiratory patterns for one minute.

\subsection{Experimental results}
\label{sub3.2}

The trained BI-AT-GRU model was tested by real-world data measured by the depth camera and the test set size was 605. Among them, there were 108 groups of Eupnea, Bradypnea and Tachypnea; 97 groups of Cheyne-stokes and Central-Apnea and 87 groups of Biots. Some wrong real-world respiration patterns were eliminated.

We trained BI-AT-GRU, BI-AT-LSTM, GRU and LSTM with the same training set with sample size of 120,000. The performance of these models was verified by the same test set (605 samples).  Results demonstrated in \textbf{Table \ref{table3}}. It can be seen from \textbf{Table \ref{table3}}: 1) four metrics of BI-AT-GRU are higher than other models, and models with bidirectional and attentional mechanisms perform better than their basic networks; 2) GRU based networks perform slightly better than LSTM based in our task; and 3) the classification accuracy of all networks maintain a high level, proving the feasibility of training the network with RSM data.
\begin{table}[ht]
\centering
\caption{Test results on real-world data.}
\label{table3}
\begin{tabular}{@{}ccccc@{}}
\toprule
\textbf{Model}     & \textbf{Accuracy} & \textbf{Precision} & \textbf{Recall} & \textbf{F1}     \\ \midrule
\textbf{BI-AT-GRU} & \textbf{94.5\%}   & \textbf{94.4\%}    & \textbf{95.1\%} & \textbf{94.8\%} \\
BI-AT-LSTM         & 90.1\%            & 90.1\%             & 91.9\%          & 91.0\%          \\
GRU                & 89.6\%            & 89.2\%             & 91.1\%          & 90.1\%          \\
LSTM               & 88.1\%            & 87.8\%             & 91.3\%          & 89.5\%          \\ \bottomrule
\end{tabular}
\end{table}

\begin{figure}[ht]
\centering
\includegraphics[height=5cm,width=5.1cm]{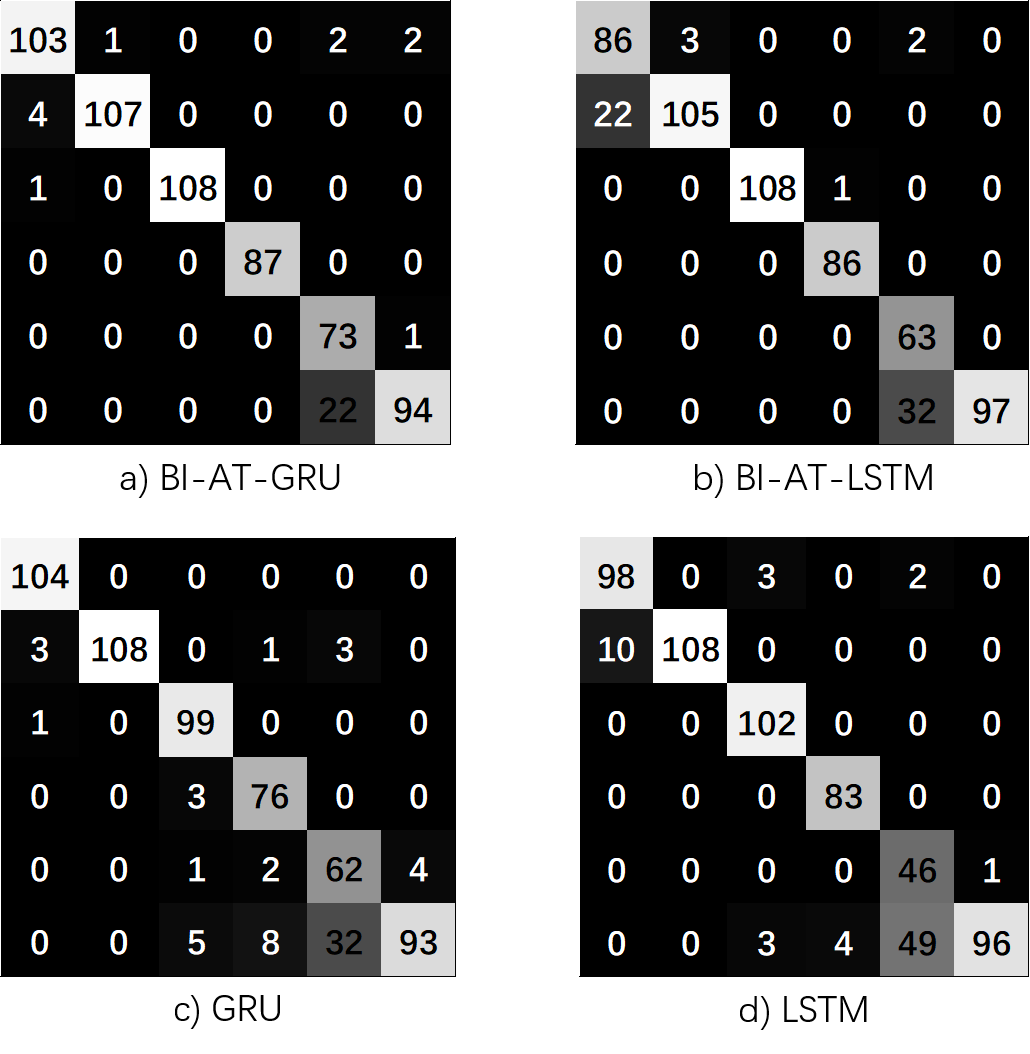}
\caption{Confusion matrix of 4 models. X axis and Y axis are the number of real labels and predicted labels respectively. From left to right or from top to bottom is: Eupnea, Bradypnea, Tachypnea, Biots, Cheyne-Stokes and Central-Apnea.}
\label{Fig6}
\end{figure}

In the confusion matrix of each model (\textbf{Fig. \ref{Fig6}}),  we can see that the classification error mainly came from prediction of Cheyne-Stokes to be Central-Apnea. The possible reason is that these patterns are only different in breathing depth, which reflected in amplitude of the waveform. While in normalization processing, amplitude of respiratory signal is sensitive to the time window. In addition, the movement of subject’s body can lead to mutations of amplitude, thus increasing the error rate. It can be seen that BI-AT-GRU has the lowest error rate, which is an important reason why it performs best. Moreover, performances of BI-AT-GRU and BI-AT-LSTM are better than their basic networks, indicating the contribution of these two mechanisms.

\section{Conclusion}
\label{sec:conclusion}
In this paper, we first apply BI-AT-GRU for classifying respiratory patterns. Because of the scarcity of real-world data, we propose a novel Respiratory Simulation Model to generate abundant training data. In validation experiments, the obtained BI-AT-GRU specific to respiratory pattern classification yields the excellent performance, and outperforms the existing state-of-the-art models. The obtained classifier has the great potential to be extended to large scale applications.

\small
\bibliographystyle{IEEEbib}
\bibliography{strings,refs}

\begin{thebibliography}{10}

\bibitem{griffiths2019guidelines}
Mark~JD Griffiths, Danny~Francis McAuley, Gavin~D Perkins, Nicholas Barrett,
  Bronagh Blackwood, Andrew Boyle, Nigel Chee, Bronwen Connolly, Paul Dark,
  Simon Finney, et~al.,
\newblock ``Guidelines on the management of acute respiratory distress
  syndrome,''
\newblock {\em BMJ Open Respiratory Research}, vol. 6, no. 1, pp. e000420,
  2019.

\bibitem{hameed2019human}
Rabab~A Hameed, Mohannad~K Sabir, Mohammed~A Fadhel, Omran Al-Shamma, and Laith
  Alzubaidi,
\newblock ``Human emotion classification based on respiration signal,''
\newblock in {\em Proceedings of the International Conference on Information
  and Communication Technology}. ACM, 2019, pp. 239--245.

\bibitem{perciavalle2017role}
Valentina Perciavalle, Marta Blandini, Paola Fecarotta, Andrea Buscemi,
  Donatella Di~Corrado, Luana Bertolo, Fulvia Fichera, and Marinella Coco,
\newblock ``The role of deep breathing on stress,''
\newblock {\em Neurological Sciences}, vol. 38, no. 3, pp. 451--458, 2017.

\bibitem{holfinger20190459}
Steven~J Holfinger, Melanie~M Lyons, Jesse~W Mindel, Peter~A Cistulli, Kate
  Sutherland, Ning-Hung Chen, Nigel McArdle, Thorarinn Gislason, Thomas Penzel,
  Fang Han, et~al.,
\newblock ``0459 diagnostic performance of symptomless obstructive sleep apnea
  prediction tools in clinical and community-based samples,''
\newblock {\em Sleep}, vol. 42, no. Supplement\_1, pp. A184--A185, 2019.

\bibitem{boulding2016dysfunctional}
Richard Boulding, Rebecca Stacey, Rob Niven, and Stephen~J Fowler,
\newblock ``Dysfunctional breathing: a review of the literature and proposal
  for classification,''
\newblock {\em European Respiratory Review}, vol. 25, no. 141, pp. 287--294,
  2016.

\bibitem{koyama2019application}
Takashi Koyama, Masanori Kobayashi, Tomohide Ichikawa, Tadamasa Wakabayashi,
  and Hidetoshi Abe,
\newblock ``An application of pacemaker respiratory monitoring system for the
  prediction of heart failure,''
\newblock {\em Respiratory medicine case reports}, vol. 26, pp. 273--275, 2019.

\bibitem{al2011respiration}
Farah~Q AL-Khalidi, Reza Saatchi, Derek Burke, H~Elphick, and Stephen Tan,
\newblock ``Respiration rate monitoring methods: A review,''
\newblock {\em Pediatric pulmonology}, vol. 46, no. 6, pp. 523--529, 2011.

\bibitem{kranjec2013novel}
Jure Kranjec, Samo Begu{\v{s}}, Janko Drnov{\v{s}}ek, and Gregor Ger{\v{s}}ak,
\newblock ``Novel methods for noncontact heart rate measurement: A feasibility
  study,''
\newblock {\em IEEE transactions on instrumentation and measurement}, vol. 63,
  no. 4, pp. 838--847, 2013.

\bibitem{al2017monitoring}
Ali Al-Naji, Kim Gibson, Sang-Heon Lee, and Javaan Chahl,
\newblock ``Monitoring of cardiorespiratory signal: Principles of remote
  measurements and review of methods,''
\newblock {\em IEEE Access}, vol. 5, pp. 15776--15790, 2017.

\bibitem{hu2018combination}
Menghan Hu, Guangtao Zhai, Duo Li, Yezhao Fan, Huiyu Duan, Wenhan Zhu, and
  Xiaokang Yang,
\newblock ``Combination of near-infrared and thermal imaging techniques for the
  remote and simultaneous measurements of breathing and heart rates under sleep
  situation,''
\newblock {\em PloS one}, vol. 13, no. 1, pp. e0190466, 2018.

\bibitem{soleimani2016remote}
Vahid Soleimani, Majid Mirmehdi, Dima Damen, James Dodd, Sion Hannuna, Charles
  Sharp, Massimo Camplani, and Jason Viner,
\newblock ``Remote, depth-based lung function assessment,''
\newblock {\em IEEE Transactions on Biomedical Engineering}, vol. 64, no. 8,
  pp. 1943--1958, 2016.

\bibitem{bae2018development}
Myungsoo Bae, Sangmin Lee, and Namkug Kim,
\newblock ``Development of a robust and cost-effective 3d respiratory motion
  monitoring system using the kinect device: Accuracy comparison with the
  conventional stereovision navigation system,''
\newblock {\em Computer methods and programs in biomedicine}, vol. 160, pp.
  25--32, 2018.

\bibitem{ernst2015respiratory}
Floris Ernst and Philipp Sa{\ss},
\newblock ``Respiratory motion tracking using microsoft’s kinect v2 camera,''
\newblock {\em Current Directions in Biomedical Engineering}, vol. 1, no. 1,
  pp. 192--195, 2015.

\bibitem{al2017real}
Ali Al-Naji, Kim Gibson, Sang-Heon Lee, and Javaan Chahl,
\newblock ``Real time apnoea monitoring of children using the microsoft kinect
  sensor: a pilot study,''
\newblock {\em Sensors}, vol. 17, no. 2, pp. 286, 2017.

\bibitem{cho2017deepbreath}
Youngjun Cho, Nadia Bianchi-Berthouze, and Simon~J Julier,
\newblock ``Deepbreath: Deep learning of breathing patterns for automatic
  stress recognition using low-cost thermal imaging in unconstrained
  settings,''
\newblock in {\em 2017 Seventh International Conference on Affective Computing
  and Intelligent Interaction (ACII)}. IEEE, 2017, pp. 456--463.

\bibitem{kim20191d}
Seong-Hoon Kim and Gi-Tae Han,
\newblock ``1d cnn based human respiration pattern recognition using ultra
  wideband radar,''
\newblock in {\em 2019 International Conference on Artificial Intelligence in
  Information and Communication (ICAIIC)}. IEEE, 2019, pp. 411--414.

\bibitem{elman1990finding}
Jeffrey~L Elman,
\newblock ``Finding structure in time,''
\newblock {\em Cognitive science}, vol. 14, no. 2, pp. 179--211, 1990.

\bibitem{hochreiter1997long}
Sepp Hochreiter and J{\"u}rgen Schmidhuber,
\newblock ``Long short-term memory,''
\newblock {\em Neural computation}, vol. 9, no. 8, pp. 1735--1780, 1997.

\bibitem{cho2014learning}
Kyunghyun Cho, Bart Van~Merri{\"e}nboer, Caglar Gulcehre, Dzmitry Bahdanau,
  Fethi Bougares, Holger Schwenk, and Yoshua Bengio,
\newblock ``Learning phrase representations using rnn encoder-decoder for
  statistical machine translation,''
\newblock {\em arXiv preprint arXiv:1406.1078}, 2014.

\bibitem{bahdanau2014neural}
Dzmitry Bahdanau, Kyunghyun Cho, and Yoshua Bengio,
\newblock ``Neural machine translation by jointly learning to align and
  translate,''
\newblock {\em arXiv preprint arXiv:1409.0473}, 2014.

\bibitem{yang2016hierarchical}
Zichao Yang, Diyi Yang, Chris Dyer, Xiaodong He, Alex Smola, and Eduard Hovy,
\newblock ``Hierarchical attention networks for document classification,''
\newblock in {\em Proceedings of the 2016 conference of the North American
  chapter of the association for computational linguistics: human language
  technologies}, 2016, pp. 1480--1489.

\bibitem{chorowski2015attention}
Jan~K Chorowski, Dzmitry Bahdanau, Dmitriy Serdyuk, Kyunghyun Cho, and Yoshua
  Bengio,
\newblock ``Attention-based models for speech recognition,''
\newblock in {\em Advances in neural information processing systems}, 2015, pp.
  577--585.

\bibitem{zhou2016attention}
Peng Zhou, Wei Shi, Jun Tian, Zhenyu Qi, Bingchen Li, Hongwei Hao, and Bo~Xu,
\newblock ``Attention-based bidirectional long short-term memory networks for
  relation classification,''
\newblock in {\em Proceedings of the 54th Annual Meeting of the Association for
  Computational Linguistics (Volume 2: Short Papers)}, 2016, pp. 207--212.

\end{thebibliography}

\end{document}